\documentclass[letterpaper]{article} 
\usepackage{aaai2027}  
\usepackage[hyphens]{url}  
\usepackage{graphicx} 
\usepackage{natbib}  
\usepackage{caption} 
\DeclareCaptionStyle{ruled}{labelfont=normalfont,labelsep=colon,strut=off} 

\usepackage{booktabs}

\usepackage{amsmath}
\usepackage{amssymb}
\usepackage{multirow}

\title{ACPO: Asymmetric Credit Policy Optimization via Mode-Local Entropy Surrogate}
\author{
    Zijun Xie\textsuperscript{\rm 1,2,$\ast\dagger$},
    Yuyang You\textsuperscript{\rm 1,3,$\ast$},
    Yongzhi Li\textsuperscript{\rm 3},
    Enlei Gong\textsuperscript{\rm 2},
    Quan Chen\textsuperscript{\rm 3},
    Yanhua Cheng\textsuperscript{\rm 3},\\
    Peng Jiang\textsuperscript{\rm 3},
    Binbin Zheng\textsuperscript{\rm 2},
    Xiaolong Liu\textsuperscript{\rm 3},
    Zeyu Chen\textsuperscript{\rm 2,$\ddagger$},
    Yadong Mu\textsuperscript{\rm 1,$\ddagger$}
}

\affiliations{
    \textsuperscript{\rm 1}Peking University \qquad
    \textsuperscript{\rm 2}Baidu Inc.\qquad
    \textsuperscript{\rm 3}Kuaishou Technology
}

\begin{document}

\maketitle

\begingroup
\renewcommand{\thefootnote}{$\ast$}
\footnotetext{Equal contribution.}
\renewcommand{\thefootnote}{$\dagger$}
\footnotetext{This work was done during an internship at Baidu.}
\renewcommand{\thefootnote}{$\ddagger$}
\footnotetext{Corresponding author.}
\endgroup

\begin{abstract}
Outcome-supervised reinforcement learning improves verifiable reasoning, but trajectory-level rewards assign the same outcome signal to all generated tokens despite their unequal contributions. Entropy provides a natural signal for token-level credit assignment, yet full-vocabulary entropy is sensitive to long-tail probabilities and induces non-local gradients. Moreover, uncertainty has different roles on positive- and non-positive-advantage trajectories. We propose Asymmetric Credit Policy Optimization (ACPO), which replaces global entropy with the complement of the top-token probability as a mode-local proxy. ACPO emphasizes uncertain decisions on positive-advantage trajectories while assigning larger negative credit to confident decisions on non-positive-advantage trajectories. It also detaches proxy gradients when sampled tokens differ from the policy mode, preventing off-target updates. We derive entropy-envelope bounds and characterize the resulting local gradient, showing advantage-aligned updates under proximal mode-preserving conditions. Experiments on mathematical and coding reasoning benchmarks, including AIME 2025 and HumanEval Pro, show that ACPO improves over entropy-aware and outcome-supervised RL baselines.
\end{abstract}

\begin{figure*}[t]
    \centering
    \includegraphics[width=\textwidth]{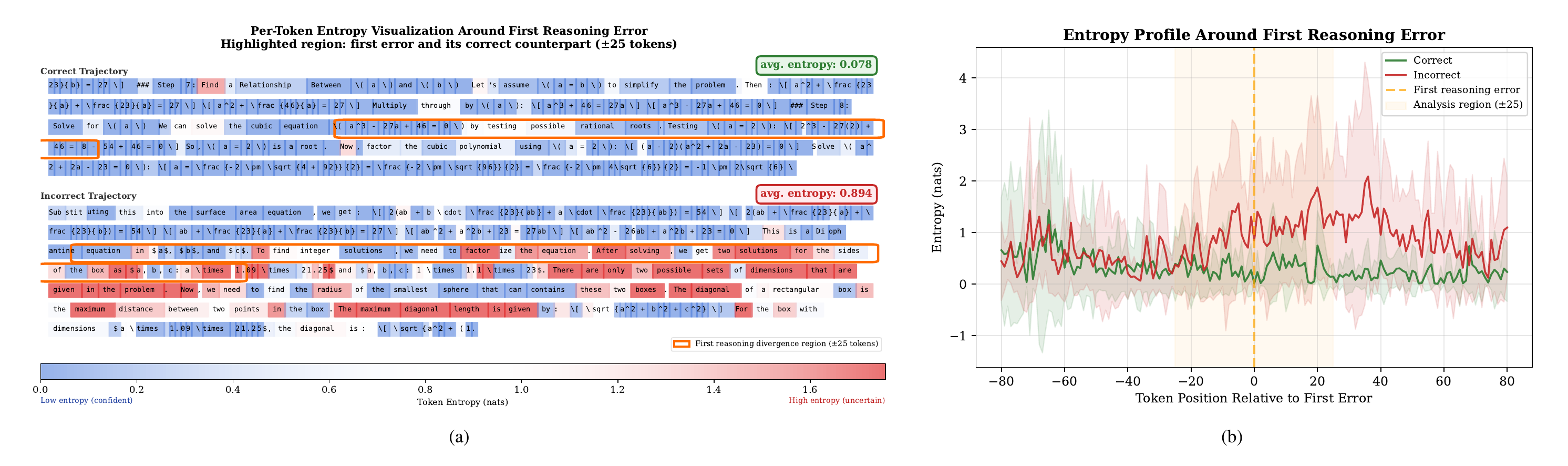}
    \caption{\textbf{Entropy around the first reasoning error.}
    (\textbf{a}) Representative correct--incorrect pair.
    (\textbf{b}) Mean entropy across aligned pairs, centered at the annotated error.
    Incorrect trajectories show elevated post-error entropy, whereas correct trajectories remain comparatively stable.
    Shaded regions denote mean $\pm$ std.}
    \label{fig:ent_comparison}
\end{figure*}

\section{Introduction}
\label{sec:introduction}

Large language model (LLM) reasoning increasingly relies on reinforcement learning with verifiable rewards (RLVR), especially for mathematics and code~\citep{Deepseek-R1, Qwen2.5-Math-7B, Qwen3-8B-Base, Qwen2, Llama3, Verl}. Unlike supervised fine-tuning, which imitates demonstrations~\citep{SFT, InstructGPT}, RLVR improves through sampled exploration and objective feedback from verifiers or compilers~\citep{STaR, DeepSeekMath, RL-survey, GSM8K}. GRPO makes this process practical by replacing the value network with group-relative advantage estimates~\citep{DeepSeekMath}. Its efficiency does not resolve credit assignment, however: a long chain-of-thought trajectory still receives a single trajectory-level outcome signal, which is then uniformly applied to tokens with very different roles in the reasoning process~\citep{CoT, SPO, PSRNSR}.

Dense process reward models and token-level critics can identify individual reasoning errors, but rely on costly step-level annotations or separate value networks that introduce substantial compute and memory overheads~\citep{PPO, PRM, PRIME}. A complementary line instead redistributes outcome credit algorithmically. DAPO and SAPO regulate policy updates through asymmetric clipping or soft gating~\citep{DAPO, SAPO, ASPO, CISPO, BAPO}, while entropy-aware methods use predictive uncertainty to locate potentially important decisions~\citep{20-80, Lp-Reg, A3PO}. The latter often discretize tokens into hard uncertainty buckets or use Shannon entropy directly as a weight~\citep{STEER, 20-80}.

Shannon entropy presents two distinct problems in this role. If differentiated, it introduces a full-distribution gradient that need not align with the sampled-token update~\citep{ProRL}. If detached, that gradient disappears, but the scalar value still depends on how residual probability is spread across a large vocabulary. The first problem concerns the backward path; the second concerns whether the forward value represents hesitation among plausible decisions. Treating them separately motivates both a mode-local proxy and explicit gradient routing.
To address these challenges, we propose \textbf{Asymmetric Credit Policy Optimization (ACPO)}. Our contributions are as follows:

\begin{itemize}
\item \textbf{Asymmetric credit assignment.}
ACPO treats uncertainty asymmetrically across advantage signs, emphasizing uncertain tokens on positive-advantage trajectories while concentrating negative credit on confident tokens.

\item \textbf{Mode-local surrogate entropy.}
We replace full-vocabulary entropy with $1-\pi_\theta(v^*)$, the probability mass outside the dominant mode, and derive deterministic bounds relating this proxy to Shannon entropy.

\item \textbf{Routed optimization.}
We introduce mismatch routing that detaches proxy multipliers for mode-mismatched tokens, preventing proxy gradients through unsampled modes. A leading-order analysis shows local advantage alignment under proximal, mode-preserving updates.
\end{itemize}

\section{Related Work}
\label{sec:related_work}

\subsection{RLHF to RLVR}

Traditional RLHF typically relies on PPO~\citep{InstructGPT, PPO} or critic-free preference learning methods such as DPO~\citep{DPO}. However, aligning reasoning models increasingly favors objective, verifiable outcome supervision. Central to this shift is Group Relative Policy Optimization (GRPO)~\citep{DeepSeekMath, guo2025deepseek}, which replaces the costly value network with group-relative mean rewards to enable efficient iterative training. Recent variants further stabilize this paradigm: DAPO~\citep{DAPO} employs decoupled clipping for large-scale updates, while SAPO~\citep{SAPO} introduces a soft, temperature-controlled gating mechanism to handle off-policy data. Our work builds directly upon the SAPO formulation, extending its sequence-level gating with intrinsic uncertainty signals for fine-grained, token-level credit assignment.

\subsection{Token-Level Credit}

A persistent challenge in reasoning tasks is the ``credit assignment'' problem---determining which specific steps in a reasoning chain led to the final outcome~\citep{SPO, RL-survey}.

\paragraph{Explicit supervision with PRMs.}
Process Reward Models (PRMs) provide dense feedback by scoring individual reasoning steps and can outperform Outcome Reward Models (ORMs) in reasoning tasks~\citep{PRM, PRIME}. However, training PRMs often requires expensive dense human annotation or complex automated verifiers~\citep{Lightman2023LetsVS, Cobbe2021Verifiers, Wang2024MathShepherd}.

\paragraph{Implicit algorithmic supervision.}
To obtain finer-grained learning signals without dense annotations,
recent methods modify token-level optimization directly. CISPO~\citep{CISPO}
and BAPO~\citep{BAPO} regulate token updates through revised importance
weighting and adaptive clipping. Related approaches use
uncertainty-guided tree sampling~\citep{Treepo} or geometric aggregation
of importance-weighted token rewards~\citep{GMPO} to improve exploration
and suppress unstable updates.

\subsection{Entropy-Aware RL}

Entropy has been widely used in RL-based language model alignment. One line of work uses entropy as an observational signal. Methods such as the ``80/20 rule''~\citep{20-80}, Archer~\citep{Lp-Reg}, A3PO~\citep{A3PO}, and GTPO~\citep{gtpo} leverage policy entropy to identify uncertain tokens, reweight advantages, or filter completions. While effective, these methods typically treat entropy as a detached heuristic rather than a differentiable optimization objective.

Another line of work directly adds entropy regularization to the policy objective, as in PPO-style entropy bonuses~\citep{PPO}. In autoregressive LLMs, this entropy is defined over the next-token distribution at each decoding step. However, it is typically aggregated via a global coefficient, serving as a generic exploration regularizer rather than a reward-aware, token-level credit signal. Moreover, directly optimizing true entropy can introduce non-local gradient components that need not align with sampled-token policy updates, as we detail in Section~\ref{sec:mode_local_surrogate}.

ACPO combines aspects of these two directions by using token-level uncertainty for fine-grained credit assignment while replacing full-vocabulary entropy with a mode-local surrogate. This design enables asymmetric updates that emphasize uncertain tokens in positive-advantage trajectories and penalize overconfident tokens in non-positive-advantage trajectories.

\begin{figure*}[t]
    \centering
    \includegraphics[width=\linewidth]{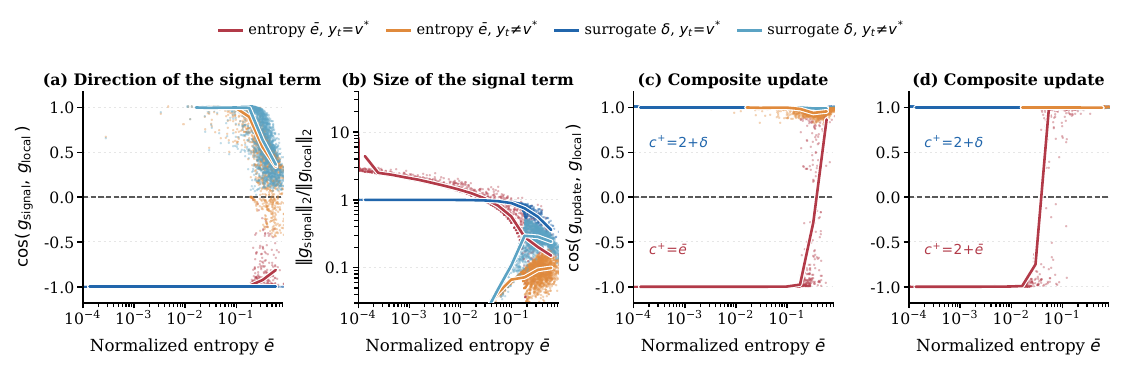}
    \caption{\textbf{Logit-space gradient geometry of normalized Shannon
    entropy and the mode-local surrogate.}
    Here, $\bar e=H/H_{\max}$ with fixed $H_{\max}=3.8545$.
    Panels (\textbf{a},\textbf{b}) show the cosine and norm ratio between
    $g_{\mathrm{signal}}$ and $g_{\mathrm{local}}$.
    For aligned tokens, the surrogate is exactly anti-parallel with norm
    ratio at most one, whereas normalized entropy can exceed one.
    Panels (\textbf{c},\textbf{d}) show the composite--local cosine for
    $\widehat A_i>0$: (c) uses $c^+=\bar e$ and (d) uses
    $c^+=2+\bar e$, both compared with $c^+=2+\delta$.}
    \label{fig:surrogate_vs_true_analysis}
\end{figure*}

\section{Preliminaries}
\label{sec:preliminary}

\paragraph{Notation.}
Let $\pi_\theta$ be a stochastic policy that maps a query (prompt) $\mathbf{q}$ to a response $\mathbf{y}$. The likelihood of generating a response $\mathbf{y}$ of length $|\mathbf{y}|$ is given by:
\begin{equation}
\label{eq:response_likelihood}
    \pi_{\theta}(\mathbf{y} | \mathbf{q}) = \prod_{t=1}^{|\mathbf{y}|} \pi_{\theta}(y_t | \mathbf{q}, \mathbf{y}_{<t}) .
\end{equation}
Each pair $(\mathbf{q}, \mathbf{y})$ receives a scalar reward $r(\mathbf{q}, \mathbf{y})$. For each query, we generate $G$ response samples and let $\mathcal{R}_{\mathbf{q}} = \{r(\mathbf{q}, \mathbf{y}_1), \dots, r(\mathbf{q}, \mathbf{y}_G)\}$ denote the set of their corresponding rewards. For brevity, we write $\pi_\theta(v) := \pi_\theta(v \mid \mathbf{q}, \mathbf{y}_{i,<t})$ when the conditioning context is clear.

\paragraph{Entropy.}
For the $i$-th response $\mathbf{y}_i$, the uncertainty of the policy at each decoding step $t$ is quantified by the Shannon entropy of the conditional distribution over the vocabulary $\mathcal{V}$. Formally, the step-wise entropy $H_{i,t}(\theta)$ is defined as:
\begin{equation}
\label{eq:ent_def}
    H_{i,t}(\theta) = -\sum_{v \in \mathcal{V}} \pi_{\theta}(v) \log \pi_{\theta}(v) .
\end{equation}
The gradient of the entropy is:
\begin{equation}
\label{eq:ent_grad}
\nabla_\theta H_{i,t}(\theta) = - \sum_{v\in\mathcal{V}} \nabla_\theta \pi_\theta(v) \log \pi_\theta(v),
\end{equation}
where the intermediate $+1$ term from the derivative of $x\log x$ vanishes because probabilities sum to one, i.e., $\sum_{v\in\mathcal{V}}\nabla_{\theta} \pi_\theta(v) = \nabla_{\theta}(1) = 0$.

\paragraph{Soft Adaptive Policy Optimization (SAPO).}
SAPO refines GSPO's~\citep{GSPO} sequence-level importance weights by replacing hard clipping with a smooth gating mechanism, making the objective sequence-coherent while retaining token-level adaptivity:
\begin{equation}
\label{eq:sapo}
\mathcal{J}_\text{SAPO}(\theta)
=\mathbb{E}
\left[
\frac{1}{G} \sum_{i=1}^{G} \frac{1}{|\mathbf{y}_i|} \sum_{t=1}^{|\mathbf{y}_i|}
    f_{i,t}\big(w_{i,t}(\theta)\big)\, \widehat{A}_{i,t}
\right],
\end{equation}
where the expectation is taken over $\mathbf{q} \sim \mathcal{D}$ and $\{\mathbf{y}_i\}_{i=1}^G \sim \pi_{\theta_\text{old}}(\cdot | \mathbf{q})$, and
\begin{equation}
\label{eq:importance_weight}
w_{i,t}(\theta) =
\frac{\pi_\theta(y_{i,t} \mid \mathbf{q}, \mathbf{y}_{i,<t})}
{\pi_{\theta_\text{old}}(y_{i,t} \mid \mathbf{q}, \mathbf{y}_{i,<t})},
\end{equation}
denotes the step-level importance sampling weight. Without a separate value function, the advantage estimate $\widehat{A}_{i,t}$ is typically sequence-level, i.e., $\widehat{A}_{i,t} = \widehat{A}_i$, and standardized within the group:
\begin{equation}
\label{eq:adv}
\widehat{A}_{i,t}=\widehat{A}_i=
\frac{r(\mathbf{q},\mathbf{y}_i)-\operatorname{mean}(\mathcal{R}_{\mathbf{q}})}
{\operatorname{std}(\mathcal{R}_{\mathbf{q}})+ \epsilon},
\end{equation}
where $\epsilon$ is a small constant for numerical stability. Using the standard sigmoid function $\sigma(\cdot)$, the soft gating function $f_{i,t}(\cdot)$ relies on an asymmetric temperature $\tau_{i,t}$ determined by the sign of the advantage:
\begin{equation}
\label{eq:soft_gate}
f_{i,t}(x) = \frac{4}{\tau_{i,t}}\,\sigma\left( \tau_{i,t} (x - 1) \right),
\;\;
\tau_{i,t} =
\begin{cases}
\tau_{\text{pos}}, & \widehat{A}_{i,t} > 0, \\
\tau_{\text{neg}}, & \text{otherwise}
\end{cases}.
\end{equation}

\section{Method}
\label{section:4}

\subsection{Entropy After Reasoning Errors}
\label{section:4.1}

Outcome-based RLVR applies one trajectory-level signal to every generated token, although tokens need not contribute equally to the outcome. To examine this mismatch, we pair correct and incorrect trajectories from the same AIME24/AIME25 prompt. For each incorrect trajectory, we annotate the first substantive reasoning error and align it with the semantically corresponding interval in the paired correct trajectory, setting the divergence point to $t=0$. Appendix~\ref{appendix:annotation_protocol} describes the annotation protocol.

Figure~\ref{fig:ent_comparison}(a) shows a representative pair: the incorrect solution introduces an ungrounded variable at the divergence point and is followed by a sustained entropy increase, whereas the correct solution remains comparatively confident. Aggregated across pairs, Figure~\ref{fig:ent_comparison}(b) shows similar entropy before the annotated error and a clear separation afterward, with incorrect trajectories developing higher-entropy suffixes.

This pattern qualifies the common interpretation of high entropy as a marker of critical decisions. On successful trajectories, high entropy can indicate hesitation among plausible alternatives, as assumed by prior entropy-aware methods such as the 80/20 rule. On failed trajectories, however, high entropy may arise after the first error as generation moves off course. Uniformly penalizing such tokens may therefore concentrate negative credit on downstream uncertainty rather than on the decisions that initiated failure. This motivates asymmetric credit assignment: ACPO emphasizes uncertain tokens on positive-advantage trajectories while concentrating negative credit on confident tokens on non-positive-advantage trajectories.

\subsection{Mode-Local Entropy Surrogate}
\label{sec:mode_local_surrogate}

A differentiable uncertainty weight should not overwhelm the
sampled-token update through its own gradient. Let $S_{i,t}(\theta)$
denote a generic uncertainty-dependent credit weight; ACPO specifies
its concrete form in Section~\ref{sec:acpo_framework}.

Omitting the advantage, which is constant with respect to $\theta$,
define
\begin{align}
u_{\mathrm{local}}
&\triangleq
\nabla_\theta\log\pi_\theta(y_{i,t}), \notag\\
u_{\mathrm{signal}}
&\triangleq
\nabla_\theta S_{i,t}.
\end{align}
The product rule then gives
\begin{align}
u_{\mathrm{update}}
&\triangleq
\nabla_\theta\!\left[f_{i,t}(w_{i,t})S_{i,t}\right] \notag\\
&=
S_{i,t}f'_{i,t}(w_{i,t})w_{i,t}u_{\mathrm{local}}
+
f_{i,t}(w_{i,t})u_{\mathrm{signal}}.
\label{eq:product_rule_breakdown}
\end{align}
Thus, differentiating the uncertainty weight adds a signal term to the
sampled-token update.

Figure~\ref{fig:surrogate_vs_true_analysis}(a,b) compares the
logit-space counterparts of these directions for the mode-local
surrogate and normalized Shannon entropy,
\[
\delta_{i,t}=1-\pi_\theta(v^*),
\qquad
\bar e_{i,t}=\frac{H_{i,t}}{H_{\max}},
\]
where $H_{\max}$ is the maximum entropy in the diagnostic batch and is
held fixed during differentiation. Appendix~\ref{gradient_diagnostic}
provides the protocol and the relation to parameter-space gradients.

On mode-aligned tokens, both signals oppose the local direction, so
the key distinction is their relative scale. For a locally fixed mode
and $y_{i,t}=v^*$,
\begin{equation}
\nabla_\theta\delta_{i,t}
=
-\pi_\theta(v^*)\nabla_\theta\log\pi_\theta(v^*)
=
-\pi_\theta(v^*)u_{\mathrm{local}}.
\label{eq:aligned_surrogate_gradient}
\end{equation}
The surrogate is therefore exactly anti-parallel to the local gradient,
with relative norm $\pi_\theta(v^*)\leq1$. Normalized Shannon entropy
has no such bound and can dominate the local term in high-confidence
regions.

Figure~\ref{fig:surrogate_vs_true_analysis}(c,d) examines positive-branch
composite directions. Panel (c) compares entropy weighting
$c^+=\bar e$ with $c^+=2+\delta$; panel (d) controls for the offset
using $c^+=2+\bar e$. Both entropy variants can overturn the local
direction on aligned, high-confidence tokens, whereas $c^+=2+\delta$
remains aligned. Thus, the difference persists after controlling for
normalization and offset.

Normalization controls the overall entropy scale but not its
dependence on residual probability allocation. Detachment removes the
entropy-gradient term but retains this forward-value sensitivity. The
mode-local surrogate instead depends only on mode confidence, bounds
its aligned gradient, and admits the entropy envelope derived in
Section~\ref{section:5.1}.

\begin{figure}[t]
    \centering
    \includegraphics[width=\linewidth]{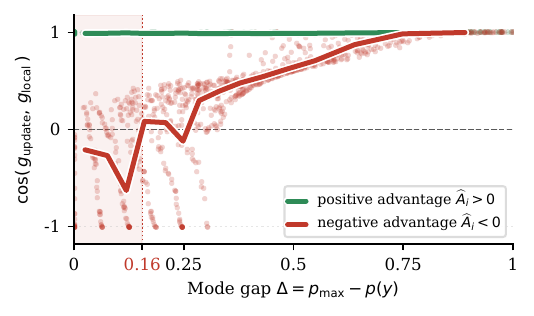}
    \caption{\textbf{Logit-space directional agreement under mode
    mismatch.} Cosine similarity between the composite and local
    directions across mode gaps $\Delta=p_{\max}-p(y)$. Positive-advantage
    updates remain aligned, whereas unrouted non-positive updates can
    reverse at small mode gaps, motivating mismatch routing.}
    \label{fig:routing_motivation}
\end{figure}

Mode mismatch creates a separate issue. When $y_{i,t}\neq v^*$, the
surrogate gradient passes through the unsampled mode and is no longer
collinear with the sampled-token gradient. Its interaction with the
local term then becomes branch dependent: it remains compatible with
the positive branch but can oppose sampled-token suppression in the
non-positive branch, producing the directional conflict in
Figure~\ref{fig:routing_motivation}. This motivates the mismatch
routing introduced in Section~\ref{sec:acpo_framework}.

\subsection{The ACPO Framework}
\label{sec:acpo_framework}

Building on the asymmetric credit motivation in
Section~\ref{section:4.1} and the proxy-gradient analysis in
Section~\ref{sec:mode_local_surrogate}, ACPO assigns token-wise credit
according to advantage sign and mode alignment. The advantage sign
selects between uncertainty-weighted positive credit and
confidence-weighted non-positive credit, while mode alignment controls
whether the proxy gradient is retained or detached.

For a sampled token $y_{i,t}$ with mode-local proxy $\delta_{i,t}$, we
define the credit multipliers $(c^+_{i,t},c^-_{i,t})$ as
\begin{equation}
\label{eq:acpo_weights}
(c^+_{i,t},c^-_{i,t})=
\begin{cases}
\left(
2+\delta_{i,t},\;
1-\delta_{i,t}
\right),
& y_{i,t}=v^*_{i,t}, \\[6pt]
\left(
3\operatorname{sg}[\delta_{i,t}],\;
3\bigl(1-\operatorname{sg}[\delta_{i,t}]\bigr)
\right),
& y_{i,t}\neq v^*_{i,t}.
\end{cases}
\end{equation}
For aligned tokens, the proxy remains differentiable. For mismatched
tokens, stop-gradient removes the off-target mode signal, while the
factor $3$ preserves the aligned proximal scales derived in
Section~\ref{section:5.2}.
Here, $\operatorname{sg}[\cdot]$ preserves its argument's forward value
while zeroing its derivative. The ACPO token-level advantage
$\widehat{A}^{\mathrm{ACPO}}_{i,t}$ is obtained by routing the
sequence-level advantage $\widehat{A}_i$ through these asymmetric
weights:
\begin{equation}
\label{eq:acpo_adv}
\widehat{A}^{\mathrm{ACPO}}_{i,t}
=
\begin{cases}
c^+_{i,t}\widehat{A}_i, & \widehat{A}_i>0, \\
c^-_{i,t}\widehat{A}_i, & \widehat{A}_i\leq0.
\end{cases}
\end{equation}
The final ACPO objective replaces the uniform advantage in SAPO
(Eq.~\ref{eq:sapo}) with this token-wise modulated advantage:
\begin{equation}
\label{eq:acpo_obj}
\mathcal{J}_{\mathrm{ACPO}}(\theta)
=
\mathbb{E}
\left[
\frac{1}{G}\sum_{i=1}^{G}
\frac{1}{|\mathbf{y}_i|}
\sum_{t=1}^{|\mathbf{y}_i|}
f_{i,t}\!\left(w_{i,t}(\theta)\right)
\widehat{A}^{\mathrm{ACPO}}_{i,t}
\right],
\end{equation}
where the expectation is taken over
$\mathbf{q}\sim\mathcal{D}$ and
$\{\mathbf{y}_i\}\sim
\pi_{\theta_{\mathrm{old}}}(\cdot\mid\mathbf{q})$.

\paragraph{Positive branch.}
For aligned positive-advantage tokens
($y_{i,t}=v^*_{i,t}$ and $\widehat{A}_i>0$), ACPO uses
$c^+_{i,t}=2+\delta_{i,t}$. The offset $2$ compensates for the
opposing proxy-gradient term introduced by the product rule. At the
proximal point, it yields the leading-order factor
$3\delta_{i,t}$, whereas $1+\delta_{i,t}$ would yield
$3\delta_{i,t}-1$ and can reverse positive updates in confident
regions. Section~\ref{section:5.2} and
Appendix~\ref{appendix:grad_derivation} provide the derivation, and
Section~\ref{sec:ablation} evaluates the offset empirically.

\paragraph{Negative branch.}
For aligned non-positive-advantage tokens
($y_{i,t}=v^*_{i,t}$ and $\widehat{A}_i\leq0$), ACPO uses
$c^-_{i,t}=1-\delta_{i,t}=\pi_\theta(v^*_{i,t})$. Its leading-order
factor is $3(1-\delta_{i,t})$, concentrating negative updates on
confident failures while downweighting uncertain downstream aftermath.
The confidence term $1-\delta_{i,t}\in[0,1]$ is bounded and introduces
no additional scaling hyperparameter.

\paragraph{Misaligned tokens.}
For mode-misaligned tokens ($y_{i,t}\neq v^*_{i,t}$),
differentiating the multiplier introduces an off-target signal through
the unsampled mode. This signal supports the positive local update but
opposes sampled-token suppression in the non-positive branch.
Figure~\ref{fig:routing_motivation} shows that this conflict is most
pronounced at small mode gaps, where the unrouted composite direction
can reverse.

The aligned analysis in Section~\ref{section:5.2} yields the proximal
factors $3\delta_{i,t}$ for the positive branch and
$\left(1+2/\tau_{\mathrm{neg}}\right)(1-\delta_{i,t})$ for the
non-positive branch, the latter being approximately
$3(1-\delta_{i,t})$ under our configuration. ACPO therefore uses the
scale-matched stop-gradient weights in
Eq.~\ref{eq:acpo_weights} for misaligned tokens; Appendix~
\ref{appendix:grad_derivation} provides the full derivation. This
removes the off-target signal while retaining the aligned
leading-order scale. The routed composite gradient is then a positive
scalar multiple of the local gradient in both logit and parameter
space and therefore preserves its direction.

\begin{figure*}[!t]
    \centering
    \includegraphics[width=\textwidth]{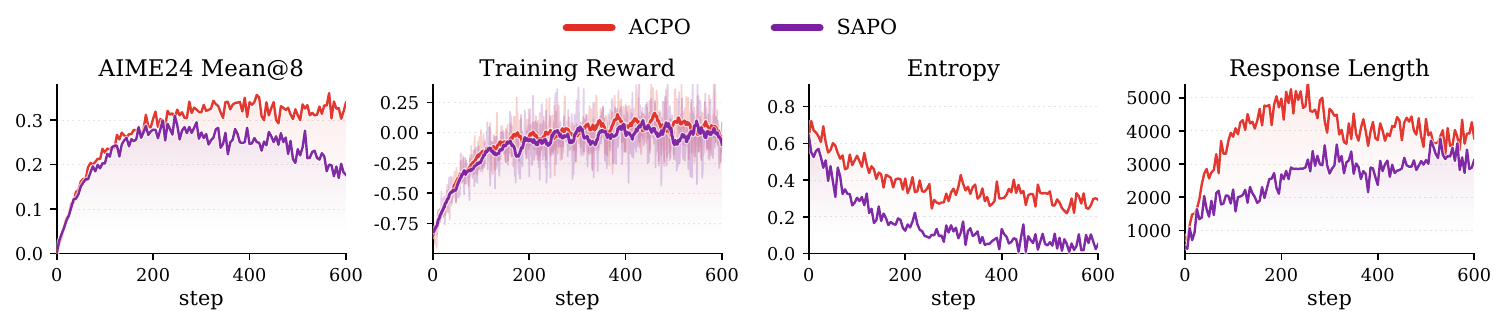}
    \caption{Training dynamics on Qwen3-8B-Base. ACPO achieves significantly higher AIME24 accuracy than SAPO while maintaining stable exploration entropy and longer response lengths.}
    \label{fig:Training}
\end{figure*}

\section{Theoretical Analysis}

\subsection{Entropy Envelope}
\label{section:5.1}

For brevity, let
$\pi_\theta(v):=\pi_\theta(v\mid\mathbf{q},\mathbf{y}_{i,<t})$,
$v^*=\arg\max_v\pi_\theta(v)$, and
$\delta=1-\pi_\theta(v^*)$. Since $v^*$ is the policy mode,
$\delta\in[0,(|\mathcal{V}|-1)/|\mathcal{V}|]$. The Shannon entropy
satisfies the deterministic bounds
$L(\delta)\leq H_{i,t}(\theta)\leq U(\delta)$, where
\begin{equation}
    L(\delta) \triangleq -\log(1-\delta),
    \label{eq:lower_bound}
\end{equation}
\begin{equation}
    U(\delta) \triangleq -(1-\delta)\log(1-\delta)
    -\delta\log\left(\cfrac{\delta}{|\mathcal{V}|-1}\right).
    \label{eq:upper_bound}
\end{equation}
The lower bound follows from the min-entropy inequality
$H_{i,t}\geq-\log\pi_\theta(v^*)$, while the upper bound is attained
when the residual mass $\delta$ is uniform over the remaining tokens.
Both bounds increase monotonically with $\delta$, so the proxy defines
a monotone envelope for Shannon entropy.

The envelope width
\[
g(\delta)
\triangleq U(\delta)-L(\delta)
=
\delta\log(|\mathcal{V}|-1)
+\delta\log\frac{1-\delta}{\delta}
\]
measures how much Shannon entropy can vary at a fixed value of
$\delta$. It is strictly concave, vanishes at both feasible endpoints,
and has a unique interior maximum. Specifically,
$\max_\delta g(\delta)=\rho^*$, where $\rho^*>0$ satisfies
\[
\rho^*+\log\rho^*
=
\log(|\mathcal{V}|-1)-1.
\]
Appendix~\ref{app:approximation_error} provides the complete
derivation.

For a distribution supported on exactly $k$ tokens, the same result
holds with $|\mathcal{V}|$ replaced by $k$; the maximum width is
$1.10$ nats for $k=10$ and $0.72$ nats for $k=5$. For practical
distributions with nonzero probability outside an effective top-$k$
set, these values are descriptive rather than deterministic, and the
full-vocabulary bounds remain valid. Figure~\ref{fig:ent-bp} provides
a complementary empirical example in which $\delta$ follows the broad
trend of normalized Shannon entropy along a representative rollout.

\subsection{Gradient Direction}
\label{section:5.2}

We next specialize the product-rule decomposition in
Eq.~\ref{eq:product_rule_breakdown} to the ACPO multipliers. As
reported in Table~\ref{table:error_analysis}, the sampled token equals
the current policy mode in $83.90\%$ of training steps. We therefore
consider the aligned case $y_{i,t}=v^*_{i,t}$ and evaluate the gradient
at the proximal limit $w_{i,t}\to1$.

The resulting leading-order policy gradient takes the form
\begin{align}
\label{eq:grad_summary_final}
\nabla_\theta \mathcal{J}_{\mathrm{ACPO}}
\propto
\mathbb{E}\left[
\nabla_\theta\log\pi_\theta(y_{i,t})
\cdot\widehat A_i
\cdot\mathcal M_{i,t}
\right],
\end{align}
where the positive and non-positive modulation factors reduce to
\begin{equation}
\label{eq:modulation_factor}
\mathcal M^+_{i,t}
=
3\delta_{i,t}(\theta),
\qquad
\mathcal M^-_{i,t}
=
3\left(1-\delta_{i,t}(\theta)\right).
\end{equation}

Equation~\ref{eq:modulation_factor} shows that positive updates scale
with the mode-probability deficit, assigning larger updates to
uncertain decisions. Conversely, negative updates scale with mode
confidence, concentrating penalties on confident decisions while
downweighting uncertain downstream regions. Since both factors are
positive, the advantage sign determines the update direction in the
aligned proximal regime. Globally detaching the multipliers removes
the signal term in Eq.~\ref{eq:product_rule_breakdown}, leaving only
forward reweighting of the local gradient.

For misaligned tokens ($y_{i,t}\neq v^*_{i,t}$),
Eq.~\ref{eq:acpo_weights} uses the stop-gradient counterparts of
Eq.~\ref{eq:modulation_factor},
$3\operatorname{sg}[\delta_{i,t}]$ and
$3(1-\operatorname{sg}[\delta_{i,t}])$. This removes the proxy
derivative while matching the leading-order scale of aligned tokens.
The resulting update is therefore a non-negative scalar multiple of
the local update and cannot undergo direction reversal.

This characterization is local to the proximal, mode-preserving
regime. Appendix~\ref{appendix:grad_derivation} provides the complete
product-rule derivation.

\begin{table*}[!t]
    \centering
    \small
    \setlength{\tabcolsep}{4pt}
        \begin{tabular}{@{}llccccccc@{}}
            \toprule
            \textbf{Model} & \textbf{Method} & \textbf{AIME24} & \textbf{AIME25} & \textbf{MATH500} & \textbf{HMMT25} & \textbf{BeyondAIME} & \shortstack{\textbf{HumanEval Pro}} & \textbf{Average} \\
            \midrule
            \multirow{7}{*}{\shortstack[l]{Qwen2.5-\\Math-7B}}
            & GRPO & $26.25$ & $16.71$ & $71.87$ & $5.20$ & $4.00$ & $42.59$ & $27.76$ \\
            & DAPO & $27.70$ & $19.56$ & $75.00$ & $6.67$ & $4.50$  & $43.29$ &  $29.44$ \\
            & 80/20 & $29.25$ & $20.50$ & $74.75$ & $7.70$ & $5.25$  & $41.46$ &  $29.81$ \\
            & GTPO & $33.97$ & $22.77$ & $75.50$ & $8.70$ & $7.12$  & $43.29$ &  $31.89$ \\
            & SAPO & $32.29$ & $21.25$ & $76.40$ & $8.64$ & $6.75$  & $43.90$ &  $31.53$ \\
            & ACPO w/o routing & $36.56$ & $25.69$ & $77.80$ & $9.86$ & $\mathbf{8.00}$ & $46.34$ &  $34.04$ \\
            & \textbf{ACPO} & $\mathbf{37.28}$ & $\mathbf{28.23}$ & $\mathbf{78.20}$ & $\mathbf{10.50}$ & $\mathbf{8.00}$ & $\mathbf{48.67}$ & $\mathbf{35.15}$ \\
            \midrule
            \multirow{7}{*}{\shortstack[l]{Qwen3-8B-Base}}
            & GRPO & $22.60$ & $16.67$ & $72.80$ & $6.67$ & $9.50$  & $35.38$ &    $27.27$ \\
            & DAPO & $24.89$ & $22.92$ & $76.80$ & $7.12$ & $11.75$   & $39.06$ &    $30.42$ \\
            & 80/20 & $26.02$ & $24.50$ & $78.22$ & $7.91$ & $12.00$   & $37.21$ &    $30.97$ \\
            & GTPO & $27.21$ & $25.45$ & $81.62$ & $8.83$ & $12.62$   & $38.49$ &    $32.37$ \\
            & SAPO & $29.06$ & $26.43$ & $82.40$ & $10.00$ & $12.25$   & $40.85$ &    $33.49$ \\
            & ACPO w/o routing & $30.31$ & $\mathbf{30.00}$ & $83.60$ & $11.04$ & $\mathbf{15.50}$   & $41.46$ &    $35.31$ \\
            & \textbf{ACPO} & $\mathbf{34.28}$ & $29.45$ & $\mathbf{84.30}$ & $\mathbf{12.89}$ & $14.37$ & $\mathbf{43.09}$ & $\mathbf{36.40}$ \\ 
            \bottomrule
        \end{tabular}
    \caption{\textbf{Main results.} Mean@8 on mathematical benchmarks and pass@1 on HumanEval Pro. Bold text indicates the best performance for each base model.}
    \label{tab:main_results}
\end{table*}

\begin{figure*}[t]
    \centering
    \begin{minipage}[b]{0.485\textwidth}
        \centering
        \includegraphics[width=\linewidth]{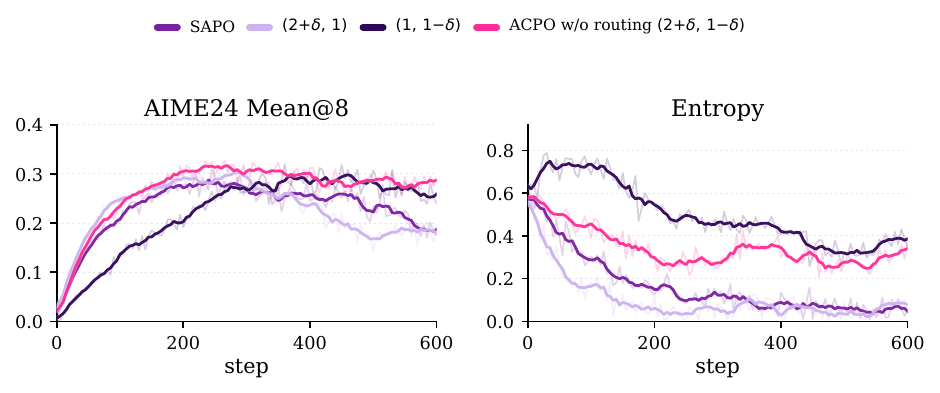}
        \vspace{2pt}
        \centerline{\small (a) Credit weight composition}
    \end{minipage}%
    \hfill%
    \begin{minipage}[b]{0.485\textwidth}
        \centering
        \includegraphics[width=\linewidth]{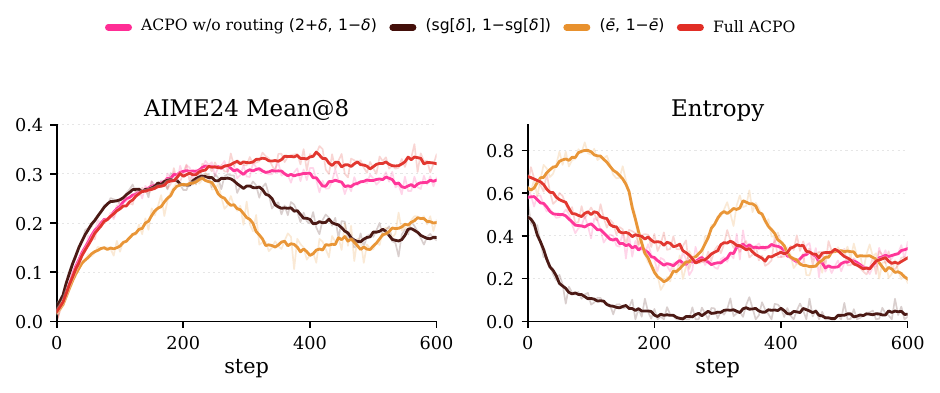}
        \vspace{2pt}
        \centerline{\small {(b) Proxy choice and gradient routing}}
    \end{minipage}
    \caption{(a) Evaluates credit weight composition and asymmetric synergy. (b) Evaluates the uncertainty proxy, product-rule gradient modulation, and the mismatch routing safeguard.}
    \label{fig:ablation_trajectories}
\end{figure*}

\section{Experiments}
\label{sec:experiments}

\subsection{Experimental Setup}
\label{sec:experimental_setup}

\paragraph{Models and training.}
We evaluate ACPO on Qwen2.5-Math-7B~\citep{yang2024qwen25mathtechnicalreportmathematical}
and Qwen3-8B-Base~\citep{qwen3technicalreport}. All methods are trained
on DAPO-Math-17k~\citep{DAPO}. We sample $G=8$ responses per prompt
with a global batch size of $128$, a mini-batch size of $16$, and a
maximum response length of $8192$ tokens. ACPO uses
$\tau_{\mathrm{pos}}=1.0$, $\tau_{\mathrm{neg}}=1.05$, and an actor
learning rate of $1\times10^{-6}$. All experiments use the same
training data, sampling configuration, and compute budget on
$8$ NVIDIA H200 GPUs.

\paragraph{Baselines.}
We compare ACPO with outcome-supervised RLVR methods, including
GRPO~\citep{DeepSeekMath}, DAPO~\citep{DAPO}, and
SAPO~\citep{SAPO}, as well as entropy-aware methods, including
80/20~\citep{20-80} and GTPO~\citep{gtpo}. Unless otherwise stated,
all methods share the same rollout and evaluation settings.

\paragraph{Evaluation.}
We evaluate mathematical reasoning on AIME24~\citep{aime24},
AIME25~\citep{aime25}, MATH500~\citep{Math500},
HMMT25~\citep{hmmt}, and BeyondAIME~\citep{bytedance_seed_2025_beyondaime},
and code reasoning on HumanEval Pro~\citep{yu2024humaneval}. We report
mean@8 accuracy on the mathematical benchmarks and pass@1 on
HumanEval Pro.

\subsection{Main Results}
\label{sec:main_results}

Table~\ref{tab:main_results} shows that ACPO achieves the highest
average performance on both base models, improving over the strongest
non-ACPO baseline by $3.26$ points on Qwen2.5-Math-7B and $2.91$
points on Qwen3-8B-Base. It also outperforms all non-ACPO baselines
across all six benchmarks for both models. The gains over SAPO are
larger on the AIME benchmarks than on MATH500, consistent with ACPO
being particularly useful for more difficult reasoning problems.
Improvements on HumanEval Pro further suggest that the learned credit
allocation transfers beyond the mathematical training domain.

Routing improves the average score by approximately $1.1$ points on
both base models. Although ACPO w/o routing performs better on two
individual Qwen3 benchmarks, the full method provides the strongest
overall performance, indicating that routing is beneficial on average
rather than uniformly across every task.

Figure~\ref{fig:Training} compares the training dynamics of ACPO and
SAPO. The two methods improve similarly during early training, after
which ACPO maintains its validation accuracy while SAPO gradually
degrades, despite broadly similar training-reward trajectories. ACPO
also maintains a moderate entropy level, whereas SAPO approaches a
near-deterministic policy. Its responses remain longer throughout
training. Together, these dynamics are consistent with ACPO avoiding
premature policy concentration; the ablations below examine which
components produce this behavior.

\begin{table}[t]
\centering
\resizebox{\columnwidth}{!}{%
\setlength{\tabcolsep}{4pt}
\renewcommand{\arraystretch}{1.1}
\begin{tabular}{@{}llr@{}}
\toprule
\textbf{Variant} & \textbf{Weights $(c^+,c^-)$} & \textbf{Mean@8} \\
\midrule
\multicolumn{3}{@{}l}{\textit{Credit composition}} \\
SAPO
& $(1,1)$
& 29.06 \\
Fixed-weight control
& $(2,1)$
& 26.12 \\
Positive modulation only
& $(2+\delta,1)$
& 28.58 \\
Non-positive modulation only
& $(1,1-\delta)$
& 27.65 \\
\textbf{ACPO w/o routing}
& $(2+\delta,1-\delta)$
& \textbf{30.31} \\
\midrule
\multicolumn{3}{@{}l}{\textit{Proxy choice}} \\
Shannon weighting
& $(\bar{e},1-\bar{e})$
& 26.68 \\
Shannon weighting + offset
& $(2+\bar{e},1-\bar{e})$
& 25.19 \\
\midrule
\multicolumn{3}{@{}l}{\textit{Gradient treatment and routing}} \\
Global SG
& $(\operatorname{sg}[\delta],1-\operatorname{sg}[\delta])$
& 27.30 \\
Global SG + offset
& $(2+\operatorname{sg}[\delta],1-\operatorname{sg}[\delta])$
& 24.75 \\
\textbf{Full ACPO}
& Eq.~\ref{eq:acpo_weights}
& \textbf{34.28} \\
\bottomrule
\end{tabular}%
}
\caption{\textbf{Ablation controls on AIME24 with Qwen3-8B-Base.}
Here, $\bar e$ denotes the normalized Shannon entropy defined in
Section~\ref{sec:mode_local_surrogate}. Unless $\operatorname{sg}$ is
shown, the uncertainty signal remains differentiable.}
\label{tab:ablation_master}
\end{table}

\subsection{Ablation Studies}
\label{sec:ablation}

We conduct ablations on Qwen3-8B-Base under the main experimental
setup, reporting mean@8 accuracy on AIME24 and policy entropy during
training. Figure~\ref{fig:ablation_trajectories} shows the training
dynamics, and Table~\ref{tab:ablation_master} summarizes the final
results.

\paragraph{Credit weight composition.}
Figure~\ref{fig:ablation_trajectories}(a) and
Table~\ref{tab:ablation_master} isolate the two credit branches. The
positive-modulation variant $(2+\delta,1)$ learns rapidly but exhibits
early entropy collapse and subsequent accuracy degradation. The
non-positive-modulation variant $(1,1-\delta)$ converges more gradually
while maintaining higher entropy and more stable late-stage
performance. Combining both branches retains fast learning while
avoiding the collapse of the positive-only variant, supporting their
complementary roles.

The fixed control $(2,1)$ underperforms, showing that uniform
amplification of positive updates does not explain the gain.
Differentiating $(2+\delta,1-\delta)$ instead yields the aligned
proximal factors $3\delta$ and $3(1-\delta)$. The product-rule
interaction thus gives both branches a common factor while assigning
positive and negative credit by uncertainty and confidence,
respectively.

\paragraph{Proxy choice.}
Since uncertainty weights act through the product rule, we compare
proxies through their induced gradients rather than only their forward
values. Figure~\ref{fig:ablation_trajectories}(b) evaluates the
differentiable Shannon weighting $(\bar e,1-\bar e)$ as a balanced
full-entropy counterpart. Its less stable entropy dynamics and weaker
late-stage accuracy are consistent with Shannon entropy's
tail-sensitive value and full-distribution gradient. The offset
variant $(2+\bar e,1-\bar e)$ also underperforms in
Table~\ref{tab:ablation_master}, supporting the mode-local proxy beyond
a particular choice of offset.

\paragraph{Proxy gradient and mismatch routing.}
The detached control
$(\operatorname{sg}[\delta],1-\operatorname{sg}[\delta])$ preserves,
up to a common factor, the aligned allocation
$3\delta:3(1-\delta)$ while removing the proxy-gradient path. Its
lower performance shows that forward reweighting alone does not
recover the behavior of the differentiable proxy.

Adding the positive offset after detachment, as in Global SG + offset,
further degrades performance because it shifts the aggregate balance
between positive and non-positive updates rather than only
redistributing credit within each branch. ACPO instead routes
misaligned tokens with the scale-matched weights
$3\operatorname{sg}[\delta]$ and
$3(1-\operatorname{sg}[\delta])$. This removes the off-target gradient
responsible for the reversals in
Figure~\ref{fig:routing_motivation} while preserving the intended local
credit scale. Full ACPO's improvement over the unrouted variant
supports the practical value of resolving this conflict.

\section{Conclusion}
\label{sec:conclusion}

We introduced Asymmetric Credit Policy Optimization (ACPO), a
token-level credit assignment method for reinforcement learning with
verifiable rewards. ACPO replaces Shannon entropy with a mode-local
surrogate, reducing tail sensitivity while bounding its aligned
gradient. Motivated by elevated entropy after reasoning errors, it
emphasizes uncertain tokens on positive-advantage trajectories and
confident tokens on non-positive ones. Mismatch routing further removes
off-target proxy signals while preserving the local update direction.
We derive deterministic entropy bounds and characterize the resulting
gradient under proximal, mode-preserving conditions. Experiments across
mathematical and coding benchmarks show consistent gains over strong
RLVR baselines, with ablations supporting the main design choices.
Future work will consider longer reasoning horizons, multi-turn
settings, and larger models.

\bibliography{references}

\clearpage
\appendix

\begin{figure*}[t]
    \centering
    \includegraphics[width=\textwidth]{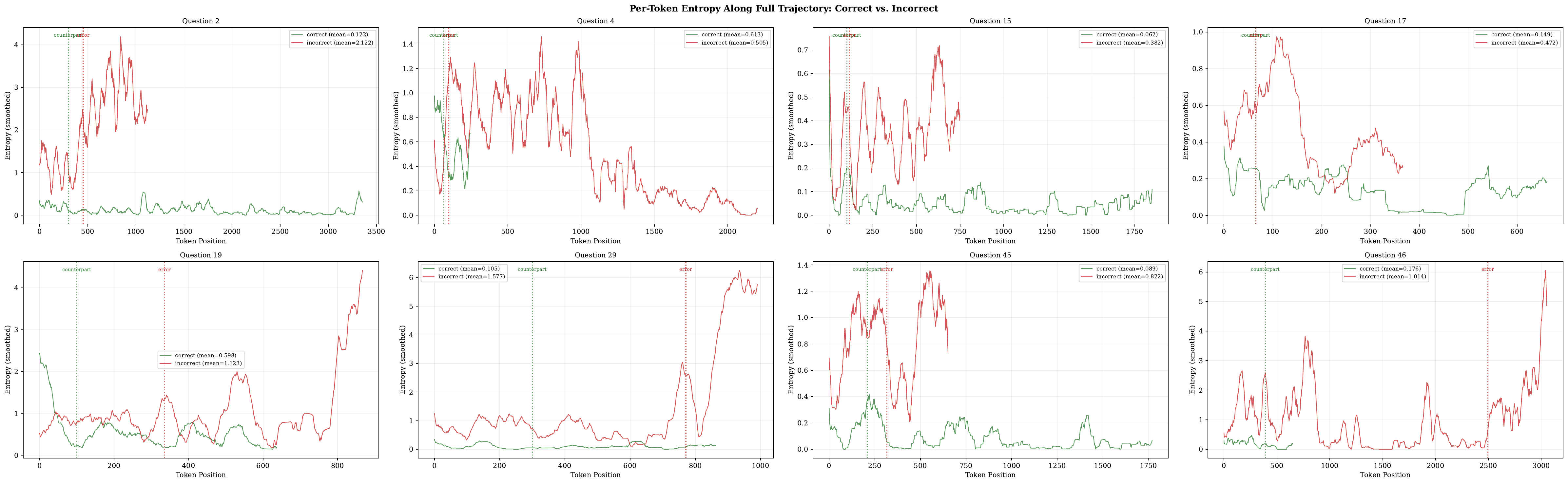}
    \caption{\textbf{Full-trajectory entropy profiles.} Smoothed token entropy for paired correct (green) and incorrect (red) trajectories from 8 AIME prompts. Dotted lines mark the annotated first error and the corresponding position in the correct trajectory.}
    \label{fig:full_trajectory_grid}
\end{figure*}

\section{First-Error Analysis}
\subsection{Annotation Protocol}
\label{appendix:annotation_protocol}
For the analysis in Section~\ref{section:4.1}, we sample eight Qwen3-8B-Base trajectories for each of the 60 AIME24/AIME25 prompts, giving 480 trajectories.

We discard prompts whose samples are all correct or all incorrect, because they provide no within-prompt contrast, and failures caused only by the generation-length limit. This leaves 8 prompts and 64 trajectories. Pairing every correct trajectory with every incorrect trajectory from the same prompt yields 91 contrastive pairs. Because a trajectory may occur in several pairs, these are 91 alignments rather than 91 independent samples.

For each incorrect trajectory, the first error is the earliest substantive mathematical or logical deviation that changes the subsequent solution path; wording, formatting, and valid alternative derivations do not count. Claude Opus 4.6 and GPT-5.5 independently propose the location and rationale, after which a human reviewer verifies the annotation. We then identify the semantically corresponding interval in the paired correct trajectory and use the two intervals to align the entropy profiles.

\subsection{Full-Trajectory Profiles}
\label{appendix:full_trajectory_entropy}

Figure~\ref{fig:full_trajectory_grid} complements the aligned analysis in Section~\ref{section:4.1} with unaligned profiles for the 8 retained prompts. In these examples, correct trajectories tend to remain lower-entropy, whereas incorrect trajectories often become more volatile after the annotated error. These profiles do not localize errors by themselves; they support the narrower observation that a high-entropy suffix can be a consequence of an earlier failure and should not automatically receive the largest negative weight.

\section{Gradient Diagnostic Protocol}
\label{gradient_diagnostic}

\paragraph{Data collection.}
We sample $G=8$ responses for each of $48$ DAPO-Math-17k prompts
from Qwen3-8B-Base before training, using the decoding configuration
of the main experiments. The model remains fixed during collection,
yielding $24{,}086$ response tokens without numerical filtering:
$21{,}537$ ($89.4\%$) are mode-aligned and $2{,}549$ ($10.6\%$)
are mode-mismatched.

\paragraph{Logit-space diagnostics.}
Although Eq.~\ref{eq:product_rule_breakdown} is defined in parameter
space, the diagnostics are evaluated with respect to the token logits
$z_{i,t}$:
\[
g_{\mathrm{local}}
=
\nabla_{z_{i,t}}\log\pi_\theta(y_{i,t}),
\qquad
g_{\mathrm{signal}}(S)
=
\nabla_{z_{i,t}}S_{i,t}.
\]
For a multiplier $c_{i,t}$, the composite gradient is
\[
g_{\mathrm{update}}(c)
=
c_{i,t}f'_{i,t}(w_{i,t})w_{i,t}g_{\mathrm{local}}
+
f_{i,t}(w_{i,t})\nabla_{z_{i,t}}c_{i,t}.
\]
All quantities are computed in closed form from the token probability
vector.

\paragraph{Uncertainty signals.}
Figure~\ref{fig:surrogate_vs_true_analysis}(a,b) compares
\[
S_{i,t}\in
\left\{
\bar e_{i,t}=\frac{H_{i,t}}{H_{\max}},
\quad
\delta_{i,t}=1-\pi_\theta(v^*_{i,t})
\right\},
\]
where $H_{\max}=3.8545$ nats is the maximum entropy over the
diagnostic batch and is held fixed during differentiation. Both signals
are evaluated on the same token states, grouped by mode alignment.
The panels report
$\cos(g_{\mathrm{signal}},g_{\mathrm{local}})$ and
$\lVert g_{\mathrm{signal}}\rVert_2/
\lVert g_{\mathrm{local}}\rVert_2$.

\paragraph{Composite updates.}
Figure~\ref{fig:surrogate_vs_true_analysis}(c,d) retains
positive-advantage tokens and evaluates the composite gradients at
$w_{i,t}=1$. With $\tau_{\mathrm{pos}}=1$,
\[
\frac{f_{i,t}(1)}
{f'_{i,t}(1)w_{i,t}}
=2.
\]
Panel (c) compares $c^+_{i,t}=\bar e_{i,t}$ with
$c^+_{i,t}=2+\delta_{i,t}$; panel (d) uses
$c^+_{i,t}=2+\bar e_{i,t}$ to control for the offset.
A reversal occurs when
$\cos(g_{\mathrm{update}},g_{\mathrm{local}})<0$.
The positive advantage is omitted because it does not affect the
cosine.

\paragraph{Mode mismatch.}
Figure~\ref{fig:routing_motivation} retains mode-mismatched tokens and
evaluates the differentiable, unrouted multipliers
\[
c^+_{i,t}=2+\delta_{i,t},
\qquad
c^-_{i,t}=1-\delta_{i,t},
\]
separately for positive- and negative-advantage branches. Tokens are
indexed by
\[
\Delta_{i,t}
=
\pi_\theta(v^*_{i,t})-\pi_\theta(y_{i,t}),
\]
and the figure reports
$\cos(g_{\mathrm{update}},g_{\mathrm{local}})$.
Within each branch, the advantage is omitted because it multiplies both
directions by the same scalar. Zero-advantage tokens are excluded.

\paragraph{Relation to parameter gradients.}
The logit-space geometry transfers exactly to the per-token gradient
of the output projection. Let $z_{i,t}=Wh_{i,t}$ and let
$G(g)=gh_{i,t}^{\top}$ denote the corresponding gradient with respect
to $W$. For any two logit gradients $g_a$ and $g_b$,
\[
\cos\!\left(G(g_a),G(g_b)\right)
=
\cos(g_a,g_b),
\qquad
\frac{\lVert G(g_a)\rVert_F}{\lVert G(g_b)\rVert_F}
=
\frac{\lVert g_a\rVert_2}{\lVert g_b\rVert_2}.
\]
Thus, the reported cosines and norm ratios exactly characterize the
direct output-head gradient geometry for each token. They should not,
however, be interpreted as statistics of the aggregated full-model
gradient, since a general parameter-space Jacobian need not preserve
these quantities.

The aligned surrogate identity is stronger. When
$y_{i,t}=v^*_{i,t}$,
\[
g_{\mathrm{signal}}
=
-\pi_\theta(v^*_{i,t})g_{\mathrm{local}}
\quad\Longrightarrow\quad
u_{\mathrm{signal}}
=
-\pi_\theta(v^*_{i,t})u_{\mathrm{local}}.
\]
Hence, its exact anti-parallel direction and relative-norm bound hold
throughout parameter space. Detaching a routed multiplier likewise
removes the signal term and leaves a non-negative scalar multiple of
the local gradient in either space.

\section{Entropy Bounds}
\label{app:approximation_error}

We derive the envelope in Section~\ref{section:5.1}. Let $\pi_\theta(v):=\pi_\theta(v|\mathbf q,\mathbf y_{i,<t})$, $v^*=\arg\max_v\pi_\theta(v)$, and $\delta:=1-\pi_\theta(v^*)$.

\subsection{Lower and Upper Bounds}
\label{app:bounds}
\textbf{Lower bound.}
Because $v^*$ has maximum probability, replacing each $\log\pi_\theta(v)$ by $\log\pi_\theta(v^*)$ gives
\begin{align}
    H_{i,t}(\theta)
    &= -\sum_{v \in \mathcal{V}} \pi_{\theta}(v) \log(\pi_{\theta}(v)) \notag \\
    &\geq -\sum_{v \in \mathcal{V}} \pi_{\theta}(v) \log(\pi_{\theta}(v^*)) \notag \\
    &= -\log(\pi_{\theta}(v^*)) \sum_{v \in \mathcal{V}} \pi_{\theta}(v) \notag \\
    &= -\log(1-\delta) \triangleq L(\delta).
\end{align}

\textbf{Upper bound.}
Separating the mode and applying Jensen's inequality to the tail gives
\begin{align}
    H_{i,t}(\theta)
    &= -\pi_{\theta}(v^*) \log(\pi_{\theta}(v^*)) \notag \\
    &\quad\; - \sum_{v \neq v^*} \pi_{\theta}(v) \log(\pi_{\theta}(v)) \notag \\
    &\leq -\pi_{\theta}(v^*) \log(\pi_{\theta}(v^*)) \notag \\
    &\quad\; - (1-\pi_{\theta}(v^*)) \log\left(\cfrac{1-\pi_{\theta}(v^*)}{|\mathcal{V}|-1}\right) \notag \\
    &= -(1-\delta) \log(1-\delta) - \delta \log\left(\cfrac{\delta}{|\mathcal{V}|-1}\right) \notag \\
    &\triangleq U(\delta). \label{eq:upper_bound_def}
\end{align}

\subsection{Envelope Width}
\label{app:gap_analysis}
Let $g(\delta):=U(\delta)-L(\delta)$. Then
\begin{equation}
    g(\delta) = \delta\log(|\mathcal{V}|-1) + \delta\log\left(\cfrac{1-\delta}{\delta}\right).
\end{equation}
The first and second derivatives with respect to $\delta$ are:
\begin{align}
    g'(\delta) &= \log(|\mathcal{V}|-1) + \log\left(\cfrac{1-\delta}{\delta}\right) - \frac{1}{1-\delta}, \label{eq:g_prime} \\
    g''(\delta) &= -\frac{1}{\delta(1-\delta)^2} < 0.
\end{align}
Thus $g$ is strictly concave. It is zero at both feasible endpoints, $\delta=0$ and $\delta=(|\mathcal V|-1)/|\mathcal V|$, and reaches its maximum in the interior.

\subsection{Maximum Width}
\label{app:max_gap}
At $g'(\delta)=0$, let $t=\delta/(1-\delta)$. Equation~\ref{eq:g_prime} becomes
\begin{equation}
    \label{eq:stationary_t_app}
    \log(|\mathcal{V}|-1) - \log t = t + 1.
\end{equation}
Substituting Eq.~\ref{eq:stationary_t_app} back into the expression for $g(\delta)$:
\begin{align}
    \max_{\delta} g(\delta) &= \delta \left( \log(|\mathcal{V}|-1) - \log t \right) \notag \\
    &= \delta (t + 1) = \frac{\delta}{1-\delta} = t.
\end{align}
Hence the maximum envelope width equals the positive solution $t$ of $t+\log t=\log(|\mathcal V|-1)-1$.

\subsection{Restricted Support}
\label{app:practical_tightness}

\begin{figure}[t]
    \centering
    \includegraphics[width=\columnwidth]{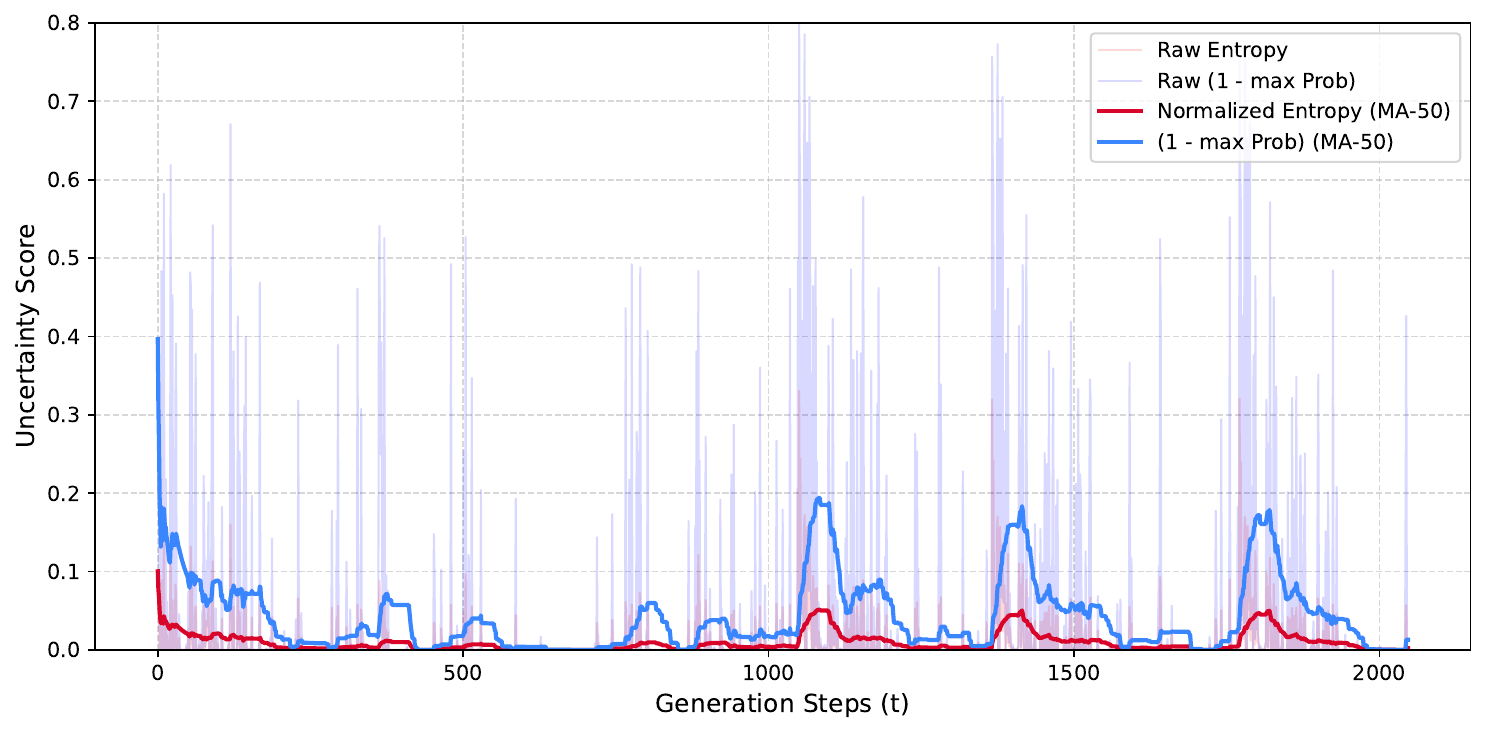}
    \caption{\textbf{Surrogate and normalized Shannon entropy.}
    On one Qwen2.5-Math-7B rollout from AIME25, the mode-local proxy follows
    the broad trend of Shannon entropy normalized by its full-vocabulary
    maximum, $H_{\mathrm{norm}}=H/\log|\mathcal V|$. Raw token-level values
    and their 50-token moving averages are shown.}
    \label{fig:ent-bp}
\end{figure}

The full-vocabulary width is a worst case. For a distribution supported on exactly $k$ tokens, the same derivation replaces $|\mathcal V|$ by $k$, and the maximum width solves $t+\log t=\log(k-1)-1$. Table~\ref{tab:error_bounds} shows that this width increases with
support size; equivalently, a smaller support yields a narrower
envelope. For an approximately top-$k$ distribution with residual mass outside the truncation, these values are descriptive rather than exact bounds; the full-vocabulary envelope remains valid. The empirical comparison in Figure~\ref{fig:ent-bp} asks the separate question of whether the mode-local proxy follows the broad entropy trend on a real rollout.

\begin{table}[t]
\centering
\small
\renewcommand{\arraystretch}{1.1}
\setlength{\tabcolsep}{4pt}
\begin{tabular}{@{}cc@{}}
    \toprule
    \textbf{Support $k$} & \textbf{Maximum envelope width} \\
    \midrule
    100 & 2.6286 \\
    40  & 1.9803 \\
    20  & 1.5235 \\
    10  & 1.1010 \\
    5   & 0.7178 \\
    \bottomrule
\end{tabular}
\caption{Maximum entropy-envelope width for an exactly $k$-supported distribution.}
\label{tab:error_bounds}
\end{table}

\section{Gradient Derivation}
\label{appendix:grad_derivation}

We derive the aligned-token modulation factors in
Eq.~\ref{eq:modulation_factor}. We omit the sequence-level advantage,
which is constant with respect to $\theta$ and is restored in
Eq.~\ref{eq:grad_summary_final}. The derivation assumes
$y_{i,t}=v^*_{i,t}$ and a local update that preserves the identity of
the policy mode.

For brevity, let
\[
p_{i,t}=\pi_\theta(y_{i,t}),
\qquad
w_{i,t}
=
\frac{p_{i,t}}{\pi_{\theta_{\mathrm{old}}}(y_{i,t})}.
\]
Under mode alignment,
$p_{i,t}=\pi_\theta(v^*_{i,t})=1-\delta_{i,t}$.

\subsection{Positive Branch}

For $\widehat A_i>0$, the single-token objective term is
\[
\mathcal J^+_{i,t}
=
f_{i,t}(w_{i,t})(2+\delta_{i,t})
=
f_{i,t}(w_{i,t})(3-p_{i,t}).
\]
Applying the product rule gives
\begin{align}
\nabla_\theta\mathcal J^+_{i,t}
&=(3-p_{i,t})w_{i,t}f'_{i,t}(w_{i,t})
\nabla_\theta\log\pi_\theta(y_{i,t})
\notag\\
&\quad
-f_{i,t}(w_{i,t})p_{i,t}
\nabla_\theta\log\pi_\theta(y_{i,t})
\notag\\
&=w_{i,t}f'_{i,t}(w_{i,t})
\nabla_\theta\log\pi_\theta(y_{i,t})
\notag\\
&\quad\times
\left[
3-p_{i,t}
-\frac{f_{i,t}(w_{i,t})}
{w_{i,t}f'_{i,t}(w_{i,t})}p_{i,t}
\right].
\label{eq:app_raw_grad_pos}
\end{align}

At the proximal limit $w_{i,t}\to1$,
$f_{i,t}(1)=2/\tau_{\mathrm{pos}}$ and
$f'_{i,t}(1)=1$. Therefore,
\[
\mathcal M^+_{i,t}
\to
3-p_{i,t}
-\frac{2}{\tau_{\mathrm{pos}}}p_{i,t}.
\]
Using $\tau_{\mathrm{pos}}=1$ gives
\begin{equation}
\mathcal M^+_{i,t}
\to
3(1-p_{i,t})
=
3\delta_{i,t}.
\end{equation}

\subsection{Negative Branch}

For $\widehat A_i\leq0$, the multiplier is
$c^-_{i,t}=1-\delta_{i,t}=p_{i,t}$, and the corresponding objective
term is
\[
\mathcal J^-_{i,t}
=
f_{i,t}(w_{i,t})p_{i,t}.
\]
Its gradient is
\begin{align}
\nabla_\theta\mathcal J^-_{i,t}
&=p_{i,t}w_{i,t}f'_{i,t}(w_{i,t})
\nabla_\theta\log\pi_\theta(y_{i,t})
\notag\\
&\quad
+f_{i,t}(w_{i,t})p_{i,t}
\nabla_\theta\log\pi_\theta(y_{i,t})
\notag\\
&=w_{i,t}f'_{i,t}(w_{i,t})
\nabla_\theta\log\pi_\theta(y_{i,t})
\notag\\
&\quad\times
p_{i,t}
\left[
1+
\frac{f_{i,t}(w_{i,t})}
{w_{i,t}f'_{i,t}(w_{i,t})}
\right].
\label{eq:app_raw_grad_neg}
\end{align}

At $w_{i,t}\to1$, the modulation factor becomes
\[
\mathcal M^-_{i,t}
\to
\left(1+\frac{2}{\tau_{\mathrm{neg}}}\right)p_{i,t}.
\]
With the configured $\tau_{\mathrm{neg}}=1.05$, the coefficient is
$1+2/1.05\approx2.905$. We round this coefficient to $3$ in the main
text to expose the symmetric leading-order form
\begin{equation}
\mathcal M^-_{i,t}
\approx
3p_{i,t}
=
3(1-\delta_{i,t}).
\end{equation}
This approximation changes only the positive scale of the negative
branch, not its dependence on mode confidence or its update direction.

Combining both branches yields the forms reported in
Eq.~\ref{eq:modulation_factor}:
\[
\mathcal M^+_{i,t}
=
3\delta_{i,t},
\qquad
\mathcal M^-_{i,t}
\approx
3(1-\delta_{i,t}).
\]

\begin{table}[t]
\centering
\small
\begin{tabular}{@{}cc@{}}
\toprule
\textbf{Probability-gap threshold} & \textbf{Cumulative coverage} \\ \midrule
$\le 1 \times 10^{-6}$ & \textbf{83.90\%} \\
$\le 0.01$ & 83.93\% \\
$\le 0.02$ & 84.07\% \\
$\le 0.05$ & 84.81\% \\
$\le 0.10$ & \textbf{86.53\%} \\
$\le 0.20$ & 89.81\% \\
$\le 0.30$ & 92.48\% \\
$\le 0.50$ & 96.13\% \\
$\le 1.00$ & 100.00\% \\ \bottomrule
\end{tabular}
\caption{Cumulative coverage of sampled-token versus mode probability gaps.}
\label{table:error_analysis}
\end{table}

\subsection{Alignment Coverage}

Table~\ref{table:error_analysis} reports the cumulative distribution of
$|\pi_\theta(y_{i,t})-\pi_\theta(v^*_{i,t})|$. The near-zero row shows
modal alignment within numerical tolerance for $83.90\%$ of the
analyzed tokens. The aligned derivation therefore applies to the
majority of tokens, while mismatch routing handles the remaining
cases.

\section{Coupling Away from the Proximal Point}
\label{appendix:dynamic_damping}

Section~\ref{section:5.2} characterizes the gradient at
$w_{i,t}=1$. For an aligned positive-advantage token, the exact
expansion in Eq.~\ref{eq:app_raw_grad_pos} can instead be written as
\begin{equation}
\begin{aligned}
\nabla_\theta\mathcal J^+_{i,t}
&=
w_{i,t}f'_{i,t}(w_{i,t})
\nabla_\theta\log\pi_\theta(y_{i,t})\\
&\quad\times
\left[
2+\delta_{i,t}(\theta)
-C(w_{i,t})
\big(1-\delta_{i,t}(\theta_{\mathrm{old}})\big)
\right],
\end{aligned}
\label{eq:app_dynamic_grad}
\end{equation}
where
\begin{equation}
C(w)
\triangleq
\frac{f_{i,t}(w)}{f'_{i,t}(w)}
=
\frac{1}
{\tau_{\mathrm{pos}}
\left[1-\sigma\!\left(\tau_{\mathrm{pos}}(w-1)\right)\right]}.
\end{equation}

The relative coefficient $C(w)$ increases with $w$. Thus, when the
current policy increases the probability of the sampled mode
($w>1$), the opposing proxy term becomes stronger relative to the
local SAPO term and can attenuate further growth. For $w<1$, its
relative influence is weaker. This does not imply gradient divergence:
the growth of $C(w)$ results from dividing by the shrinking gate
derivative, while the unfactored scalar coefficient
$f_{i,t}(w)\pi_\theta(v^*)$ remains bounded. The sign of the complete
update also depends on the rollout-policy probability, so this analysis
describes a damping tendency rather than a stability guarantee.

\end{document}